\documentclass[letterpaper]{article}
\usepackage{natbib,alifeconf}

\title{The Effect of Social Learning on Individual Learning and Evolution}
\author{Chris Marriott$^{1}$ \and Jobran Chebib$^2$ \\
\mbox{}\\
$^1$University of Washington, Tacoma, WA, USA 98402 \\
$^2$University of Calgary, Calgary, AB, Canada T2N 1N4 \\
dr.chris.marriott@gmail.com}

\begin{document}
\maketitle

\begin{abstract}
We consider the effects of social learning on the individual learning and genetic evolution of a colony of artificial agents capable of genetic, individual and social modes of adaptation.  We confirm that there is strong selection pressure to acquire traits of individual learning and social learning when these are adaptive traits.  We show that selection pressure for learning of either kind can supress selection pressure for reproduction or greater fitness.  We show that social learning differs from individual learning in that it can support a second evolutionary system that is decoupled from the biological evolutionary system.  This decoupling leads to an emergent interaction where immature agents are more likely to engage in learning activities than mature agents.
\end{abstract}

\section{Introduction}
When agents possess both genetic adaptation operating on a generational time scale and learning operating on the agent's time scale there is the potential for interaction between these adaptive mechanisms. \cite{SSE} have surveyed studies into this interaction and show conditions under which the presence of learning can accelerate or decelerate genetic adaptation.

In particular \cite{PSK} found that if the fitness function in the direction of adaptation is concave and the step size of the learning algorithm is small genetic adaptation can occur at an increased rate.  This acceleratory effect was seen in a number of different simulations (for instance \cite{HN, FM, M, L}).

On the other hand if this condition is not upheld, or indeed the opposite conditions exists, then the learning can slow the natural genetic adaptation (see \cite{P, A95,DGPR,BMR}).  Some simulations have shown both acceleratory and deceleratory effects under different conditions (\cite{A00,PKS,PSK}).  

Social learning is a form of learning that arises from social situatedness (\cite{LZ}) and is characterized by agents interacting with one another in order to learn.  Social learning can accelerate learning beyond that of individual learning strategies (see \cite{DP,AP,MPD}) and most notably is its ability to support cumulative cultural evolution (\cite{MWL}) as witnessed in human culture.  While a wide range of mechanisms of learning have been studied in conjunction with evolution we have found little evidence of the study of social learning mechanisms on evolution.  Further, since social learning is a distinct form of learning identified by its social as opposed to individual nature, it may also interact with the individual learning processes.  In this study we explore the interactions between social learning, individual learning and genetic adaptation.

We have designed a model that allows for evolution of reproductive, individual learning and social learning abilities in conjunction with traits for fitness.  These abilities are not just on/off for our agents but instead can be participated in more or less than others.  The abilities also come at a cost to the agent (in time) which limits the maximum fitness achievable by the agent.

The model we have developed allows for the genetic information and the learned information to be expressed in the same format, allowing for direct comparison between genetic adaptation and the individual and social learning processes.
\section{Experimental Setup}

\subsubsection{Learning Task}
Our simulation involves agents that gather resources from a number of different resource sites.  Each site has five locations where resources might be found.  Only one, two, or three of the locations actually have resources while the others are empty.  We call this value the reward of the site.  Say an agent is at a resource site with three resources hidden in the five locations.  The agent must select the order it will check the locations.  Once the agent has all three resources it can stop checking, and the cost of this gather attempt is the number of locations checked.  This cost is measured in time units.

Agents will have to select which sites to gather from each day and what order to check locations at those sites.  The resource sites may not be repeated in a day, but can be repeated again the next day.  The next day the resource site will have the same number of resources and they will be stored at the same locations.  This allows for the agent to adapt its strategy on future days.

More formally we can think of a resource site as a subset of $\{1,2,3,4,5\}$  with size equal to the reward.  For instance with reward equal to 3 the resource site might be the set $\{1,2,4\}$.  An attempt at this site by the agent is characterized by a permutation of the set $\{1,2,3,4,5\}$, for instance, $[5,4,2,1,3]$. The cost to the agent (in time units) at this site is equal to the number of entries in the permutation that must be processed before the subset representing the site has been covered by the elements of the permutation.   In our example it requires the first 4 elements of $[5,4,2,1,3]$ to cover the set $\{1,2,4\}$ resulting in a cost of 4 time units spent to gather 3 resources.  Notice the maximum cost is always 5 and the minimum cost is equal to the reward.

An agent is allocated 50 time units in a day and may gather from as many sites as it has time units.  Thus optimally an agent can gather 50 rewards from resource sites in one day if all time units are dedicated to gathering.  An agent may also spend its time units on activities that may increase its long term fitness.  Specifically it may spend units attempting to reproduce, learn individually, or learn socially.  These tasks each take one time unit and may be taken more than once in a day.

\subsubsection{Agent Design}

The agent consists of a genome and a memome.  The genome and the memome are internally identical, but play different roles in the agent.  The genome is inert information that remains static for the duration of the agent's life.  It is used during asexual reproduction to create a new agent. The memome is dynamic and may change over the lifetime of the agent.  It is used to select daily behavior and may be spread socially.

Genomes and memomes are internally identical and in this paragraph we will describe these internals from the perspective of a genome.  However, everything is equally true of the memome.  A genome  consists of 50 geneplexes, each describing a series of actions that could be carried out by an agent in a single day.  A geneplex's total cost is the time cost of all actions in the geneplex and is always less than or equal to 50. A geneplex's fitness is equal to the total reward that its actions accrue (ties are broken favoring geneplexes that spend less time attaining the same total reward or spend some time reproducing or learning).  When a genome is mutated the five geneplexes with greatest fitness replace the five geneplexes with least fitness and the copies are mutated.  A geneplex mutates by changing the strategy at a randomly selected resource site, adding a randomly generated action, or removing a randomly selected action.

At birth an agent will copy its genome to create its memome.  When a day's activities are to be selected the agent takes the memeplex with highest fitness and carries out its list of actions.  Its daily reward is the total number of resources these actions accrue.

A new agent is created from a parent selected randomly but not uniformly from the population.  An agent's chance of being selected is equal to the number of breeding tickets it has.  It is rewarded one breeding ticket for every five resources it gathered today and one for each time unit the agent spent on reproducing today.  So agents better at gathering and who spend more time on reproduction are more likely to be a parent.  The new agent's genome is a mutated clone of its parent's genome.

When an agent is born the agent with least fitness dies to maintain a constant population size.  An agent's fitness is equal to the average daily reward during its life minus its age.  As a result agents that die are either poorer at gathering, older, or both.

During a day an agent may spend time learning.  For each occurrence of an individual learning action in its daily action list the agent will mutate its memome once.  This activity will change the memome but leaves the genome unchanged.  The new memome is used to select the next day's actions.

An agent is selected for participation in a social gathering with probability proportional to the number of time units spent on socially learning.  The selected agents then contribute their best five memeplexes into a pool.  The best five in the pool are then redistributed to the agents replacing their five least fit memeplexes.  Again this exchange affects only the memome leaving the genome unchanged.

\subsubsection{Population Design}

We considered three isolated populations of 50 agents.   Each agent in the initial population has a genome consisting of geneplexes with only a single random gathering action.  No genome was seeded with reproducing, individual learning, or social learning actions.  

The first population served as a control and agents were not capable of learning in any way.  Their only means of behavioral adaptation is genetic evolution.  Any actions spent by these agents on learning were wasted actions as they had no result and thus these actions were slightly maladaptive for these agents.  The memome and genome were identical in these agents since their memome was incapable of change.

Agents in the second population were capable of learning from the environment but not from each other.  Any actions spent on learning from others were wasted actions in this population and thus were slightly maladaptive.  Agents had to evolve to learn before they would begin learning.

Agents in the third population were capable of all the learning strategies, and like the agents in the second population, they needed to evolve to learn individually or socially.  These learning capabilities were separate and thus evolving one did not imply evolving the other.

We gathered data on the maximum fitness of agents in each population over time.  This allowed us to track the optimization occurring in each population.  We also tracked the maximum fitness of geneplexes in the genomes of agents in each population.  This allowed us to isolate and track genetic optimization.  We also gathered data on the frequency of actions for reproducing, individual learning, and social learning among our agents.  We selected an epoch length of 10000 days with data samples every 20 days.  We have averaged results over 159 runs for presentation.  With such a large number of runs the confidence intervals are very small and are omitted for clarity.

\section{Observations and Discussion}

\begin{figure*}[!t]
\begin{center}
\includegraphics[width = 7in]{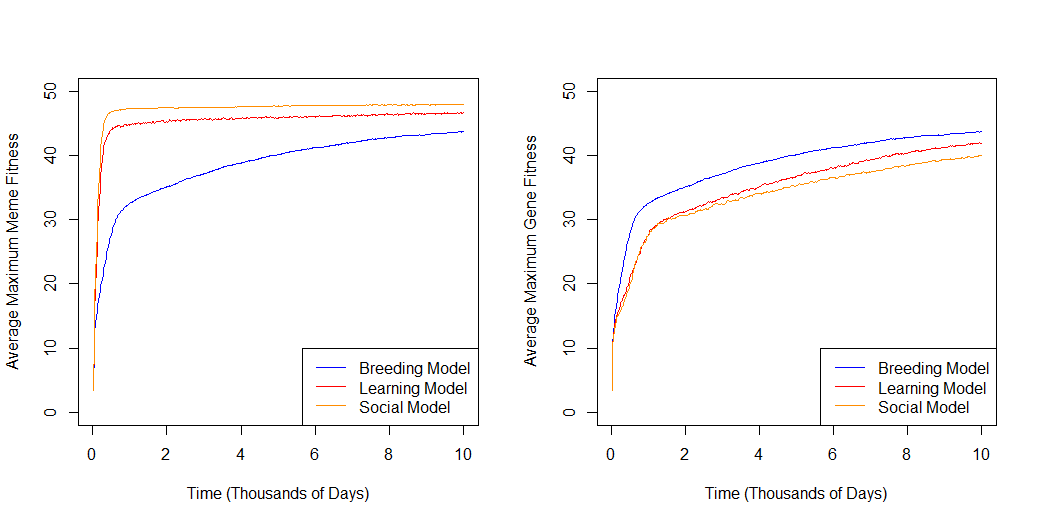}
\caption{Average maximum fitness of memes (left) and genes (right) in breeders, learners and socializers.}
\label{fig1}
\end{center}
\end{figure*}

\subsubsection{Overall Fitness}
By day 1500 the average maximum memetic fitness of agents in the social learning group increases to a maximum of 48 (see Figure~\ref{fig1}).  Agents cannot both attain the optimum and maintain their social learning behavior.  In most runs agents converged on a shared memeplex with one social learning action and one individual learning action (and occasionally a reproduction action).

This behavior can be considered optimal in that a memeplex of higher fitness would have to replace either the social learning action or the individual learning action with a gathering action, and thus lose the benefit these actions give.  Replacing the social learning action might make the agent more fit, but would prevent that action from being shared with others and the more fit memeplex would die with the agent.  Replacing the individual learning action would stop individual optimization of the memeplex making it very unlikely to increase fitness.  Unlikely as this event is in our model we do observe this occurring infrequently.

By day 1500 the average maximum fitness of agents from the individual learning group has increased to 45.  By the end of the 10000 day epoch this has been further optimized to nearly 47.  

At the 10000 day epoch individual learners have an average of four actions allocated to reproducing and learning in the memome.  Individual learners are responsible for optimizing their own memeplexes whereas social learners can rely on the community to communicate the optimums to them.  As a result the individual learners must maintain greater commitment to learning even when near the optimum fitness to be competitive.  

The average maximum fitness of agents from the non-learning group increases to only 34 by day 1500.  At the end of the 10000 day epoch the average fitness is nearly 44.  This is still a few steps from the optimal and this is because we see at the end of the epoch the maximal memeplex had on average four reproduction actions (and one of the null actions).

This behavior should be seen as nearly optimal.  Breeding tickets are awarded one for every five rewards achieved and one for every reproduction action.  This would suggest that there is an equilibrium between selection pressures rewarding greater fitness and those rewarding time spent on reproduction.  Since on average five time units are spent on non-gathering actions the fitness of these agents can theoretically be increased by five.  However this would award the agents only a single additional breeding ticket (in exchange for four lost), greatly decreasing their odds of reproducing.  On the other hand agents that spend much more time on breeding actions than four (or five) will reduce their fitness making them more vulnerable to early death due to low fitness.  The midpoint between these opposing selective forces appears to be at five time units.

These average maximum fitness results support a common hypothesis regarding the rate of optimization utilizing these different modes of learning (\cite{DP, AP, MPD}).  Specifically, genetic optimization is seen as slower than optimization through individual learning or learning from the social group.  Furthermore, learning from the social group can maintain learned strategies in the memosphere and thus carry out cummulative learning.  This leads to the hypothesis that social learning is the most rapid mode of optimization.  

\subsubsection{Genetic Fitness}

Our interest extends further to the effect that these modes of learning have on the genetics of the agents using them.  Modern hypotheses (\cite{SSE}) suggest that the learning mechanisms can both heighten selection, thus accelerating genetic adaptation, and shield it, thus decelerating genetic adaptation.

Both the individual learners and the social learners had less fit genomes compared to the control non-learning group for most of the simulation.  At day 1500 both learning groups have a average maximum genome fitness of 30 while the non-learning groups was 34 (see Figure~\ref{fig1}).  This indicates that the selection pressure operating on the genomes according to fitness levels was shielded by the learning of both kinds in the early days of the simulation.  This is predicted in instances like ours where the learning can operate to flatten the fitness landscape.

Optimization is shielded for fitness levels in the range 15 to 30 during an "easy task" of growing the initially short geneplexes to ones that use all 50 time units.  After this point the optimization problem increases in difficulty since the geneplexes then can only be optimized by improving the strategies of individual gathering attempts. 

There does not appear to be any further shielding effect on the social learners during the "difficult task".  That is the slope of improvement for non-learners and social learners is very close after day 1500 (both improve 10 fitness in 8500 days).  In contrast the individual learners appear to optimize at a more rapid rate (12 fitness in 8500 days).  

In individual learners differences between agents' average daily fitnesses becomes dependent on early performance  and speed of convergence while solving the "difficult task".  Agents that have less fit genomes perform worse early on and those that take longer to converge spend more time at lower reward levels.  Thus, conditions seem met for an acceleration of adaptation of both the genes for learning and the genes for more efficient gathering.  

That is, agents that have a more fit genome or occupy positions that allow them to more rapidly overcome genome deficiencies are selected for in the population.  On the one hand this pressure rewards agents with more fit genomes because they both perform better early on and can optimize more rapidly from this better position.  On the other hand this pressure would favor agents that can learn rapidly indicating selection for individual learning actions (see discussion below).  This seems to be a case of Baldwin's initial hypothesis where the selection pressure accelerates adaptation of plasticity simultaneously with greater fitness.

Noting this we might expect also that social learners should benefit from this same acceleration though we did not find this in our data.  We did find strong evidence of increased selection for social learning actions (see discussion below) in these agents and we hypothesize the benefits of these actions greatly outweigh benefits to early performance.

\begin{figure*}[!t]
\begin{center}
\includegraphics[width = 7in, height = 6in]{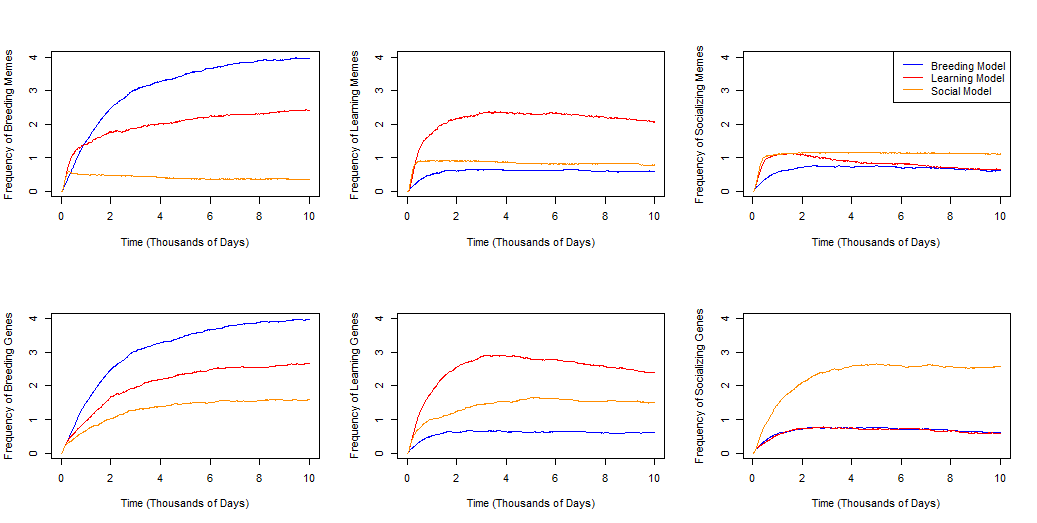}
\caption{Average number of breeding, learning and socializing actions in the genomes and the memomes of breeders, learners and socializers.}
\label{fig2}
\end{center}
\end{figure*}

\subsubsection{Reproductive Actions}

Non-learning agents converged on an average of four reproduction actions by the end of the epoch.  This is due to the selection pressure that is applied related to the increased chance of reproduction for these actions.  We expect the same selection pressure to exist for the learning agents as well and we see evidence for that in the genomes of the learners (see Figure~\ref{fig2}).  The individual learners steadily increase to an average of about 2.75 reproductive actions per agent.  Notably this average is lower than that in the non-learners indicating that this selection pressure is shielded by the learning.  We also see evidence of this shielding effect in the social learners who increase to an average of about 1.6 reproductive actions per agent. 

A contributing factor to the shielding effect is that the memome is used to determine the number of breeding tickets issued, not the genome.   As a result the number of reproductive actions in the genome do not contribute long term to the number of breeding tickets issued to the agent, as the memome is assumed to change from its initial state in learners.  As such the reproductive actions in the genome contribute most to the chance of breeding early in life (before learning potentially replaces these actions).  This also means that an early genetically determined reproductive behavior can be learned away.  This provides fewer opportunities for more fecund genomes to express their greater fecundity resulting in a shielded selection pressure.

We see that in individual learners the number of reproduction actions in the memome is coupled to the number in the genome.  While prior to day 3000 there are more learning actions in the memome than the genome, this changes after about day 3000.  The reason for this change is as the genetic fitness approaches the optimum the pressure to learn decreases. 

In the social learners we see a different dynamic.  The average reproductive actions in the memome seem uncoupled from the genome.  The reproductive actions in the genome increase due to pressure (if shielded) for increased reproduction.  In the memome we have a depreciating curve, indicating selection pressure against breeding in the memome.  Breeding actions do not help spread the memeplex, but they do help spread the geneplex, and so we see divergent selection pressures here and a decoupling of the adaptive processes.

This decoupling is due to the independence of the evolution of memeplexes from the evolution of geneplexes and this independence results in cumulative cultural evolution.  The individual learning of each individual learner must begin optimizing from the initial condition coded in its genome (hence a coupling of these processes).  The social learner in contrast can begin from whatever state is provided by its community.  Thus it need not "reinvent the wheel" and repeat learning steps that others have already carried out.  This leads to a memosphere whose evolution is independent of (decoupled from) the co-evolving genosphere.  As a result the selection pressure that occurs in the genosphere for reproductive actions does not exist in the memosphere.  This may also be the case for individual learners, but only the social learners have decoupled the memosphere's evolutionary path so the selection pressure can be effective.

Nonetheless the genetic selection pressure still penetrates into the genome.  Since agents still have early performance based on their genome, early reproductive performance is still determined by the reproductive actions in the genome.  In our simulations the expected number of days before a new agent participates in its first social learning gathering is quite small.  After this point in the agent's life the selection pressure is shielded by the memome but before this point it can favor agents that have more reproductive actions in the genome.

\subsubsection{Individual Learning Actions}

Learning actions in the non-learning group were null actions and were slightly maladaptive.  The maladaptivity came from the fact that the action had a temporal cost of one time unit that could have been better spent on harvesting.  This would suggest that there would be selection against having this trait in the genome. We see that the average memome (and thus genome) in the non-learning group has around 0.5 occurrences of each null learning action or about 1 full occurrence of a null learning action (see Figure~\ref{fig2}).  In our simulation null actions and no actions are rewarded the same in fitness.

Then the frequency of null learning actions in the memome of our non-learners speaks to the average number of time units not spent on harvesting.  Specifically we'd expect an equal number of empty spaces to null learning actions (since they are equally favored by the fitness function) so an average of one null learning action suggests an average of two time units not spent on harvesting in the genome of non-learners.  These unused time units are integral to the formation of more fit variants as they are replaced with new harvesting actions.

Individual learning is adaptive for the individual learning agents.  We see a very rapid assimilation of the individual learning action into the population after its first appearance in the genosphere.  Since with our settings it takes about 100 days to replace the entire population we expect it takes about this minimum time for the individual learning action to be represented in every living agent.  That is, non-learning agents are quickly replaced by individual learning agents in the population once learning agents evolve.

This selection for individual learning continues after this initial assimilation in both the memome and the genome.  For instance we see early on that the memome's average frequency of individual learning actions exceeds that of the genome's average frequency.  The first agents that evolved to learn then quickly learned to learn more (their memome's acquired a second or third occurrence of the individual learning action to accelerate learning).  In addition, selection for learning actions in the genome continues at a rapid rate, eventually overtaking the frequency in the memome at around two individual learning actions per agent on average.

This switch is best explained with reference to when individual learning is the most advantageous to an agent.  Young agents have lower fitnesses (since their daily reward is tied to their genome) and thus to be competitive they need to rely on their individual learning actions to optimize their memome rapidly.  The more rapidly they do this the smaller overall hit they will take to their fitness.  Thus selection strongly favors agents that have more individual learning actions in their genomes.

This selection does not work similarly in the memome.  Older agents will have optimized their memome to a higher fitness and gain less benefit from individual learning.  Indeed we see that some rare individuals learn to stop learning near the end of their lives if they can swap out that action with a more fit harvesting action.  More commonly we see that agents that at birth had two or three individual learning actions in their genome will optimize at least one of those actions away by their old age.  

Another interesting trend in the individual learners is the steady decrease of the of the number of individual learning actions in both the genome and the memome after peaking around 3000 days.  This decrease suggests that as the simulation progresses the selection pressure that once favored higher numbers of individual learning actions is lessening.  This occurs as the genome optimizes and approaches the memome's fitness levels.  Since the genome represents the starting conditions of the individual learning process the more fit this initial state is the less in need of learning the agents are.  This happens both within the agent's life (making individual learning actions less common in memomes of older agents than younger) and within the larger population in generational time (making individual learning actions less common in the genomes of agents born later than those born earlier).

Individual learning is still an important skill in the social learning group.  While social learning facilitates sharing of memeplexes and the maintenance of a cross-generational culture, it has no means of optimization.  All optimization in the social learners is still due to the individual learning actions in their memomes.  We see that selection for individual learning in the genomes is rapid in the social learners.

Interestingly we also see a more rapid increase in the presence of individual learning actions in the social learners' memomes in the early stages paralleling the individual learners learning to learn more.  However, in the social learners this rapid increase is very quickly countered when the fitness of the shared memeplex reaches near optimal and a new pressure to learn less appears.  The consequence of this is that the shared memeplex of the social learners tends to have exactly one individual learning action.  Notice this is the minimum number to ensure continued optimization allowing it to satisfy the pressure for continued optimization while minimizing the cost of that optimization.  

Further in rare cases the optimization is capable of eliminating that last individual learning action from the memome since there is no need for further optimization when it is at the near optimal. The reason this is a rare occurrence is because for this to occur the agent would have to use an individual learning action to optimize its genome and then eliminate the action that did the optimization and result in a genome that is more fit.  This can only occur with a very lucky sequence of mutations in our setup and so is very rare.  Nonetheless this is the reason why the average number of individual learning actions in the memome is not exactly one but instead less than, indicating the number of runs in which the social learners were capable of achieving the lucky sequence of mutations to eliminate this action.  About one fifth to one quarter of the simulations had eliminated this action by the end of the epoch.

Again the striking difference in shapes between the curves of the individual learning actions in the memome and genome of the social learners can be attributed to a decoupling of selection pressures on the memome and the genome.  While young agents will be more likely to individually learn as soon as they socially learn their genetic proclivity is overridden by the shared memeplex's behavior.

\subsubsection{Social Learning Actions}

Finally consider the social learning actions.  We know these actions are null actions in both the non-learners and individual learners and as such we see very low frequencies in both the genomes and memomes of these agents.  For the reasons discussed above regarding tie breaks we see non-zero occurrences of these actions.

However in the social learners the social learning action is efficacious.  Further social learning is adaptive for these agents, but even more so if it is accompanied by an individual learning action.  Without the individual learning action a collection of social learners will only share the most fit memeplex in their memomes.  Since they are not engaged in individual learning this represents the most fit geneplex in their genome.  As a result agents can still benefit from social learning in the absence of individual learning, though this benefit is negligible.

There is a parallel and independent selection pressure for individual learning.  Both selection pressures are strong and so agents are expected to quickly evolve these actions, but very commonly they evolve roughly simultaneously.  The two pressures are synergistic in that agents that evolve both traits outperform those with only one.  The results of this synergistic selection is that the genomes of agents rapidly include at least one copy of each of the learning actions, with social learning actions being selected for more aggressively than individual learning actions.  

We also see that selection for the social learning action in the genome continues to increase whereas in the memome this frequency is highly stable with an average of  one per agent.  Indeed the reason that this average is greater than one is that newborn agents are frequently born with more than one social learning action in their genome (and thus their initial memome) and this slight deviation from the norm of a few agents in the social community is enough to keep this average slightly over one.  This is further evidence of the decoupling of the memosphere and the genosphere.

This decoupling leads to a very interesting dynamic.  We see that social learners optimize to have on average about 2.5 social learning actions in the genome of an agent.  A mature adult (one that has already engaged in social learning at least once) will have only one social learning action in the memome of the agent.  This means that a newborn or immature agent will be 2.5 times more likely to engage in social learning than a mature adult.

This is an important feature since this means that younger, immature social learners are more likely to engage in the activity that will make them more fit, but once they have engaged in that activity they do not attempt to perform it as frequently anymore allowing for younger agents to have the opportunity.  It is also still important for mature agents to engage in social learning but their role is as an instructor, not a learner in most of these meetings.  

The pressures that lead to this state are once again independent and decoupled.  Young agents, due to selection on early performance on their genome, face a strong selection pressure to socialize early.  This is accommodated by having as many social learning actions as possible in their genome.  Old agents, due to selection pressure related to fitness, face a strong selection pressure to socialize less to make room for more harvesting.  This is accomodated by having the minimum number of socializaing actions as possible in the memome. The equilibrium found by this interaction is to make young agents socialize as soon as they can while older agents socializing only enough to ensure that young agents learn the dominant memeplexes in the memosphere.

Once again this is clear evidence of the decoupling of the memosphere from the genosphere in the social learners.  Nonetheless, the early stages of a memome's development is still dependent on the genome of the agent.  Once the agent learns from the community this dependence ends and the systems are decoupled.

\section{Conclusions}

Our experiment shows how social learning, individual learning and genetic adaptation can co-exist in a population of agents and some of the interactive dynamics between these optimizing forces.  Specifically we see both instances of accelerated and decelerated genetic adaptation in response to these learning mechanisms in our model.

We see clear selection shielding in the early stages of the simulation as a result of both individual learning and social learning.  The cause of this shielding is that the learning mechanisms flatten the fitness function, thus equalizing selection pressure on all agents.  Later in the simulation we see slight selection boosting for the individual learners.  The cause of the boosting is selection for early performance and a Baldwin effect.

In both the learning populations we see shielding of selection pressures for reproductive actions in the genome.  These selection pressures still exist but are weaker in agents capable of individual and/or social learning.  Shielding is greatest in social learners.  This indicates that individual learning and social learning in particular can weaken or even eliminate selection pressure for genetic reproductive behavior while simultaneously suppressing the same behaviors.

In both learning populations we also see boosting of selection pressures for learning actions in a genome.  The fact that both learning mechanisms are highly adaptive leads to learners always outperforming non-learners and thus rapid assimilation of new learning actions into the genome occurs.  Further we see a trend of increased selection for learning actions in the genome even when there is contrasted selection for removing those actions from the memome.  This emerges from the opposite need of young and old agents for learning.  The young agents are necessarily less fit and so can really benefit from learning, whereas the old agents tend to be highly optimized and thus further learning is wasted effort.  

This trend appears in both individual learners and social learners but is most apparent in the social learners.  This is due the evolution of the memosphere decoupling from the evolution of the genosphere made possible by the shared memosphere.  Not only is this decoupling responsible for the rapid optimization of the social learning group but it results in divergent and uncorrelated behavior frequencies between genome and memome.  

To be clear the evolution of the memosphere and genosphere in the individual learners is independent but the starting position of both optimization procedures begin with the genome, and thus the two procedures are coupled.  In the social learners the genosphere and memosphere are coupled for a very short period before the memosphere reinitializes with the best memeplex from the community.  This allows the memosphere's evolution to begin from a different point than the genosphere's evolution.  This implies the decoupling is not $100\%$ but there is still a small link.  We say that while decoupled the memosphere is still tethered to the genosphere.

The decoupling of the two evolutionary spheres allows for opposite selection pressures to be fully efficacious in each sphere.  The best evidence of this we have is that while selection for reproductive, individual learning, and social learning actions is strong in the genosphere of social learning agents, it is weak, or even negative, in the memosphere of social learning agents.  The memosphere attempts to eliminate the extraneous reproduction actions and learning actions, while the genosphere is selecting for greater numbers of these actions in the genome.  The greater numbers in the genoshpere lead to the emergent behavior that young agents are more likely to reproduce, learn on their own, or learn from others than their older counterparts.  This is beneficial to them since they are the least fit members of the population at this time.  

\footnotesize
\bibliographystyle{apalike}
\bibliography{example}

\end{document}